\documentclass{article}
\usepackage{amsmath,amssymb,theorem}
\usepackage{color}

\theorembodyfont{\upshape}

 \makeatletter
 \@addtoreset{equation}{section}
 \makeatother

 \newtheorem{ittheorem}{Theorem}
 \newtheorem{itlemma}{Lemma}
 \newtheorem{itproposition}{Proposition}
 \newtheorem{itdefinition}{Definition}

 \newtheorem{itremark}{Remark}
 \newtheorem{itclaim}{Claim}
 \newtheorem{itcorollary}{\bf Corollary}

 \newenvironment{theorem}{\addtocounter{equation}{1}
 \begin{ittheorem}}{\end{ittheorem}}

 \newenvironment{lemma}{\addtocounter{equation}{1}
 \begin{itlemma}}{\end{itlemma}}

 \newenvironment{proposition}{\addtocounter{equation}{1}
 \begin{itproposition}}{\end{itproposition}}

 \newenvironment{definition}{\addtocounter{equation}{1}
 \begin{itdefinition}}{\end{itdefinition}}

 \newenvironment{remark}{\addtocounter{equation}{1}
 \begin{itremark}}{\end{itremark}}

 \newenvironment{claim}{\addtocounter{equation}{1}
 \begin{itclaim}}{\end{itclaim}}

 \newenvironment{proof}{\noindent {\bf Proof.\,}
 }{\hspace*{\fill}$\qed$\medskip}

 \newenvironment{corollary}{\addtocounter{equation}{1}
 \begin{itcorollary}}{\end{itcorollary}}

 \newcommand{\be}[1]{\begin{eqnarray*}\label{#1}}
 \newcommand{\ee}{\end{eqnarray*}}

 \newcommand{\bl}[1]{\begin{lemma}\label{#1}}
 \newcommand{\el}{\end{lemma}}

 \newcommand{\br}[1]{\begin{remark}\label{#1}}
 \newcommand{\er}{\end{remark}}

 \newcommand{\bt}[1]{\begin{theorem}\label{#1}}
 \newcommand{\et}{\end{theorem}}

 \newcommand{\bd}[1]{\begin{definition}\label{#1}}
 \newcommand{\ed}{\end{definition}}

 \newcommand{\bcl}[1]{\begin{claim}\label{#1}}
 \newcommand{\ecl}{\end{claim}}

 \newcommand{\bp}[1]{\begin{proposition}\label{#1}}
 \newcommand{\ep}{\end{proposition}}

 \newcommand{\bc}[1]{\begin{corollary}\label{#1}}
 \newcommand{\ec}{\end{corollary}}

 \newcommand{\bpr}{\begin{proof}}
 \newcommand{\epr}{\end{proof}}

 \newcommand{\bi}{\begin{itemize}}
 \newcommand{\ei}{\end{itemize}}

 \newcommand{\ben}{\begin{enumerate}}
 \newcommand{\een}{\end{enumerate}}

 \def \ba {\begin{array}}
 \def \ea {\end{array}}

 \def \qed {{\heartsuit\hfill}}

\def \qed {{\square\hfill}}

\let\F=E     

\def \qed {{\square\hfill}}

\def\eqref#1{(\ref{#1})}

\begin{document}

\title{Critical control of a genetic algorithm}

 \author{
Rapha\"el Cerf
\\
\\
Universit\'e Paris Sud and IUF}

\maketitle


\begin{abstract}
\noindent
Based on speculations coming from statistical mechanics and the conjectured
existence of critical states, I propose
a simple heuristic in order to control the mutation probability
and the population size of a genetic algorithm.
\noindent
\end{abstract}

\bigskip

Genetic algorithms are widely used nowadays, as well as their cousins 
evolutionary algorithms. The most cited initial references on genetic 
algorithms are the beautiful books of Holland \cite{HO}, who tried to
initiate a theoretical analysis of these processes, and of Goldberg \cite{GO},
who made a very attractive exposition of these algorithms.
The literature on genetic algorithms is now so huge that it is beyond
my ability to compile a decent reasonable review.
For years, there has been an urgent and growing demand for guidelines to
operate a genetic algorithm on a practical problem.
On the theoretical side, progress is quite slow and somehow disappointing
for practitioners.
The theoretical works often deal 
with a simple toy problem, otherwise the behavior
of the genetic algorithm is too complex to be amenable to rigorous
mathematical analysis.
Here I propose
a simple heuristic in order to control efficiently the genetic algorithm,
based 
on speculations coming from statistical mechanics and the conjectured
existence of critical states.
Although it is quite simple, I have not been able to locate this heuristic
in the literature, and I hope it will be useful. 
Apart from my own belief, it
is supported by several empirical studies, the most notable one
being the work of Ochoa \cite{OCH}, and it is in accordance
with several conclusions and ideas appearing in the 
work of 
van Nimwegen 
and 
Crutchfield 
\cite{NP3}.
\medskip

\noindent
{\bf Error threshold}.
The fundamental notion on which the heuristic is based is the notion
of error threshold, introduced by Manfred Eigen in 1971 \cite{EI1}.
Eigen analyzed a simple system of replicating molecules and demonstrated
the existence of a critical mutation rate.
Above this mutation rate, the information carried by the molecules is destroyed
by the mutations.
This fundamental result lead to the notion of
quasi\-spe\-cies developed by Eigen, McCaskill
and Schuster \cite{ECS1}, which plays an important role
in evolutionary biology.
The critical mutation rate can be explicitly computed on 
the simplest fitness landscape:
the space $\{0,1\}^n$ and the fitness function equal to $1$ everywhere
except one point where it is equal to
$\sigma>1$.
In this situation, the critical mutation rate per bit is
$$p_m\,=\,\frac{\ln\sigma}{n}\,.$$
This is the sort of result we dream of in the context of genetic algorithms: an
explicit simple expression for the critical mutation probability.
%
Several researchers have already argued that the notion of error threshold plays a role
in the dynamics of a genetic algorithm.
This is far from obvious, because Eigen's model is formulated for an infinite
population model. However there is evidence that a similar phenomenon occurs in finite
populations as well, and also in genetic algorithms.
In her PhD thesis \cite{OCH}, Ochoa demonstrated the occurrence of error thresholds in genetic
algorithms over a wide range of problems and landscapes. This very interesting
work is published in a series of conference papers.
\medskip

\noindent
{\bf Optimal population size and mutation rate}.
The dream of a practitioner is to have a
set of optimal parameters to solve his specific
problem. On a few simple examples it is possible to find empirically the optimal rates,
by sampling several runs of the genetic algorithm with different parameters.
This was done by Ochoa in her PhD thesis. She concluded that there exists
a relationship between the optimal mutation rate and the error threshold.
An important contribution
of Ochoa's work is to try to relate quantitatively the optimal mutation rate
with the error threshold.
Cervantes and Stephens investigated further this idea \cite{CS}.
One of the most interesting and inspiring 
work on the theory of genetic algorithms
I have read over the last years is the series of papers by
van Nimwegen, Crutchfield and Mitchell
\cite{NCM1,NCM2,NP1,NP2,NP3}.
In these papers, the authors perform a theoretical and experimental 
study of a genetic algorithm on a specific
class of fitness functions. Their analysis rely on techniques from mathematical
population genetics, molecular evolution theory and statistical physics.
Among the fundamental ingredients
guiding the analysis are the quasi\-species model, the
error threshold and metastability.
In the last work of the series \cite{NP3},
van Nimwegen
and 
Crutchfield 
describe an entire search effort surface
and they introduce a generalized error threshold in the space of the
population size and mutation probability delimiting a set of parameters where
the genetic algorithm proceeds efficiently.
\medskip

\noindent
{\bf Phase transitions}.
A basic goal of statistical mechanics is to understand the collective behavior
of particles governed by simple microscopic rules. 
Typically, the particles are driven by two antagonistic effects:
entropy and energy.
Interesting models present a phenomenon of phase transition: 
there exists a critical
point or a critical curve
in the parameter space separating a region where energy effects
dominate from a region where entropy effects dominate.
The system is most interesting at criticality, where both forces compete equally.
Perhaps the most studied
model is the Ising model. 
There has been remarkable progress recently in the rigorous understanding of
the critical Ising and percolation models in two dimensions \cite{WW}.
The error threshold is in fact a particular type of phase transition
\cite{EI3}. The antagonistic forces in presence
are mutation and selection and this threshold separates a regime where
selection dominates from a regime where mutation dominates.
In a genetic algorithm, the crossover operator complicates the dynamics
and either it shifts the critical points or it creates new ones.
This phenomenon has been observed independently
by Rogers, Pr\"ugel--Bennett and Jennings
\cite{RPJ} and by Nilsson Jacobi and Nordahl \cite{NN}.
%
\medskip

\noindent
{\bf Efficient search}.
An efficient search procedure should realize a delicate balance between
an exploration mechanism and a selection mechanism. 
This general idea is present 
in numerous works dealing with random optimization. 
I have believed for several years that a search procedure works best if it is close
to a critical point, which realizes an optimal balance between the exploration
and the selection mechanisms. This view is supported by the general knowledge coming
from statistical mechanics, and it is also expressed in several previous works
\cite{OCH,NP3,RPJ}.
Unfortunately, phase transitions and critical points are sharply 
defined only for infinite
systems. Moreover the computation of the critical points is very hard and complex,
it can be achieved only for specific models, like Eigen's model or the
two dimensional Ising model, and it requires great mathematical skills. For
the three dimensional Ising model, the critical point can only be estimated
numerically.
Hence the task of computing the optimal parameters of a genetic algorithm on
a specific problem is a formidable one, clearly much harder than solving the
problem itself.
\medskip

\noindent
{\bf Self--organized criticality.} 
My next hope is to try to adapt the parameters of the genetic algorithm
during the search in order to reach a critical regime. Systems which are
driven naturally towards a critical state have attracted a lot of interest
since the seminal work of Bak, Tang and Wiesenfeld \cite{BTW}. 
These systems are said to exhibit
self--organized criticality.
Several researchers have tried to incorporate such mechanisms to design
optimization procedures. An interesting example is the extremal optimization
\cite{BOE}. In his PhD thesis \cite{WHI}, Whitacre investigates the occurrence of
self--organized criticality in evolutionary algorithms.
Krink, Thomsen and Rickers used successfully mechanisms inspired by self--organized criticality to control evolutionary algorithms \cite{KTR}.
My aim here is to propose a very simple heuristic 
to achieve a critical control of a genetic algorithm.
\medskip

\noindent
{\bf Critical control of the mutation.}
When running a genetic algorithm, we are not looking for
the optimal mutation probability, rather we look for a control of
the mutation probability which allows to explore efficiently the
space. Several researchers have already worked on this idea and
proposed different possible schemes to adapt the parameters of
the genetic algorithm. A review is presented in \cite{ERZ}.
Here I propose an adaptive procedure which receives a simple feedback
from the search.
The conjectural picture I have in mind is the following.
The genetic algorithm running on a fitness landscape is
a finite population model, approximating an infinite population model.
This infinite model presents several phase transitions, depending on the
geometry of the fitness landscape.
In a way, there is a phase transition associated to each local maximum.
The parameters of a genetic algorithm should be adjusted in order to be
close to the phase transition corresponding to its current position.
When the algorithm escapes from a local maximum and finds a better point,
the parameters should be completely readjusted from scratch.
In practice, we need a simple criterion to decide whether the mutation parameter
is above or beneath the local critical value.
Considering Eigen's model of quasispecies and keeping in mind that we
are dealing with a finite population, I propose the following simple criterion.
If the best fitness observed in the population decreases, then the mutation
probability is above the critical value.
If the best fitness observed in the population is constant, 
then the mutation
probability is below the critical value.
These speculations lead to the algorithm presented on page~$7$.
As for the procedure to control the mutation, there are plenty of
choices. A very simple possibility is to use a dichotomy procedure.
We use two extremal values
$\alpha$ and
$\beta$ which bound the abstract critical mutation probability
and we do as follows.
To initialize the mutation control, we set
$\alpha=0$ and
$\beta$ to a reasonable upper bound on the critical mutation probability.
In the case of a genetic algorithm working with binary words of length $n$,
the initial value of
$\beta$ should be of order $(\ln c)/n$, where $c$ is larger
than the ratio between the maximum and the minimum 
of the fitness function. One should not take 
$\beta$ too large, otherwise the mutations will destroy all the
relevant information in the population, without hope of recovering 
the interesting points during the subsequent steps.
The initial probability mutation is then set to 
$p_m= (\alpha+\beta)/2$.
To increase the mutation probability, we set successively
$\alpha =p_m$ and
$p_m= (\alpha+\beta)/2$.
To decrease the mutation probability, we set successively
$\beta =p_m$ and
$p_m= (\alpha+\beta)/2$.
When the algorithm decreases the mutation, it is because the best
fitness has decreased, and the hope is 
to recover the best points
of the previous generations with reversed mutations. This is likely to
happen only if the mutation rate is not too high, hence it seems important
to be able to adjust adequately the initial upper bound~$\beta$.
Another possibility is to use elitism, which leads to the variant
of the previous algorithm presented on page~$8$.
The idea is natural: whenever the best fitness has decreased, we reintroduce
by force the lost best solution, and we decrease the mutation probability.
I think that this algorithm is more interesting than a standard
search procedure incorporating elitism, because it is likely to escape
from a local maximum much quicker. When exploring the landscape around a local
maximum, it is reasonable to explore incrementally the neighborhood, starting
with the points which are close in a mutational sense
to the current best solution and
proceeding then with further and further points. The progressive increase
of the mutation rate implements this strategy to some extent. 
Another good reason to increase the mutation rate is to avoid
premature convergence of the population, a problem that has been
observed since the early days of genetic algorithms. The algorithm
I propose tries to run with the largest reasonable
mutation probability, which maintains the widest diversity in the population
without losing the best fit individual.
\medskip

\noindent
{\bf Critical control of the population size}.
A very interesting conclusion of \cite{NP3} 
is the existence of a critical population size
below which it is practically impossible to reach the global optimum.
A similar conclusion was obtained in the simpler framework of generalized
simulated annealing \cite{CE1}: 
within a specific asymptotic regime of low mutations
and high selection pressure, the convergence to the global maximum could
be guaranteed only above a critical population size. 
These works
support the idea of the existence of a critical population size. 
From now on, let us suppose that this idea is correct.
If we adhere to the belief that a search procedure is more efficient
when it is running close to a critical state, we should also try to adjust 
the population size close to the critical one, in the same way 
as we did with
the mutation.
It is of course clear that we want a population size larger than the critical
one, because an algorithm with too small a population will not reach the
global maximum.
One may however wonder why it would be a problem to use a population
size much larger than the critical one. I imagine that it would lead to a waste
of computational resources. Indeed it does not make sense to use
a genetic algorithm with a large population size to find the maximum
of a concave function, because a population of a few individuals will
certainly suffice and do the job much faster.
Another reason is that it seems best to have only a very small fraction
of the population sitting on the best current points, in order to 
use most of the computational resources to explore the vicinity
of these points. This is also a very interesting conclusion of \cite{NP3}.
With this in mind, I propose the more elaborate algorithm presented
page~$9$.
\bigskip

\noindent
{\bf What to do next}.
I hope that the critical mutation control of the genetic algorithm will
lead to real practical improvement. The benefit of controlling
the population size 
is more conjectural, but I believe it is a very interesting direction
to try: if the genetic algorithm is still
stuck at a local maximum after convergence of the mutation probability,
then one should increase the population size in order
to escape and find a better solution.
A further possibility is to perform a simultaneous critical
control of the mutation probability and the population size during the
run of the algorithm. To do that will require slightly more complicated
methods, because two parameters have to be optimized simultaneously,
but this seems really to be a promising direction to explore. 
From an algorithmic point of view, it seems easier to vary first
the mutation probability and then the population size.
Another issue is the use of elitism. Elitism is a straightforward
mechanism guaranteeing the asymptotic convergence. 
Instead of using
an elitist algorithm, one can try
to exert an adequate control on the parameters of the algorithm
in order to enforce elitism.
\bigskip

\noindent
{\bf Summary of the heuristic}.
The heuristic I propose is quite simple. The parameters
of a genetic algorithm should be set close to a conjectured critical
line describing the phase transition associated with its current
localization in the fitness landscape.
In order to adjust the parameters,
we observe the behavior of the best fit individual from one 
generation to another and we proceed as follows:

\noindent
$\bullet$ If the best fitness decreases
strictly, then the genetic algorithm is operating
in the regime where mutation dominates. Thus the mutation
probability should be decreased or the population size should be increased.

\noindent
$\bullet$ If the best fitness is constant,
then the genetic algorithm is operating
in the regime where selection dominates. Thus the mutation
probability should be increased or the population size should be decreased.

\noindent
$\bullet$ If the best fitness increases strictly,
then
the genetic algorithm has successfully  escaped from a local maxima,
the mutation control should be reinitialized and the population size
should be reset to two.

\noindent
The idea of adapting the parameters of the genetic algorithm
is not a new one and it has already been seriously investigated by many
researchers~\cite{ERZ}. However the philosophy behind
the control described above is quite different from what I have seen
in the literature. For example,
a common approach is to reward operators which lead to
good solutions, with the hope that they will create even better solutions.
My suggestion comes from a different principle, which is in fact quite opposite. 
The crucial idea is
that a delicate interaction between exploration and selection is the key to
an efficient search. 
In order to create the critical equilibrium, I suggest 
to favor the mechanism which currently does not produce
good results! 

\newpage
~\vfill
\centerline{\bf Critical mutation control}
\medskip

\noindent
Initialization
\medskip

$\bullet\quad$ Initialize the population
\medskip

$\bullet\quad$ Find the best fit individual $\widehat y$
\medskip

$\bullet\quad$ Initialize the mutation control
\medskip

\noindent
Main loop
\medskip

$\bullet\quad$ Set $\widehat x=\widehat y$
\medskip

$\bullet\quad$ Apply selection
\medskip

$\bullet\quad$ Apply mutation
\medskip

$\bullet\quad$ Apply crossover
\medskip

$\bullet\quad$ Find the best fit individual $\widehat y$

\medskip
If $\text{fitness}(\widehat x)=\text{fitness}(\widehat y)$ then
\medskip

\qquad Increase mutation probability
\medskip

\qquad Goto Main loop
\medskip

If $\text{fitness}(\widehat x)>\text{fitness}(\widehat y)$ then
\medskip

\qquad Decrease mutation probability
%
\medskip

\qquad Goto Main loop
\medskip

If $\text{fitness}(\widehat x)<\text{fitness}(\widehat y)$ then
\medskip

\qquad Reinitialize the mutation control
\medskip

\qquad Goto Main loop
\bigskip
\bigskip

\noindent
\centerline{\bf Dichotomy procedure to change mutation}
\medskip

\noindent
Initialization: $\alpha=0$, $\beta=(\ln c)/n$, 
$\,p_m= (\alpha
+\beta )/2$
\medskip

\noindent
Mutation increase: $\alpha=p_m$, 
$\,p_m= (\alpha +\beta )/2$
\medskip

\noindent
Mutation decrease: $\beta=p_m$, 
$\,p_m= (\alpha +\beta )/2$
\vfill
\newpage
~\vfill

\centerline{\bf Critical mutation control with elitism}
\medskip

\noindent
Initialization
\medskip

$\bullet\quad$ Initialize the population
\medskip

$\bullet\quad$ Find the best fit individual $\widehat y$
\medskip

$\bullet\quad$ Initialize the mutation control
\medskip

\noindent
Main loop
\medskip

$\bullet\quad$ Set $\widehat x=\widehat y$
\medskip

$\bullet\quad$ Apply selection
\medskip

$\bullet\quad$ Apply mutation
\medskip

$\bullet\quad$ Apply crossover
\medskip

$\bullet\quad$ Find the best fit individual $\widehat y$

\medskip
If $\text{fitness}(\widehat x)=\text{fitness}(\widehat y)$ then
\medskip

\qquad Increase mutation probability
\medskip

\qquad Goto Main loop
\medskip

If $\text{fitness}(\widehat x)>\text{fitness}(\widehat y)$ then
\medskip

\qquad Decrease mutation probability
\medskip

\qquad Reintroduce $\widehat x$ in the population
\medskip

\qquad Set $\widehat y=\widehat x$
\medskip

\qquad Goto Main loop
\medskip

If $\text{fitness}(\widehat x)<\text{fitness}(\widehat y)$ then
\medskip

\qquad Reinitialize the mutation control
\medskip

\qquad Goto Main loop
\vfill
\newpage
\noindent

\newpage
~
\vfill
\centerline{\bf Critical mutation and size control with elitism}
\medskip

\noindent
Initialization
\medskip

$\bullet\quad$ Set the population size to two
\medskip

$\bullet\quad$ Initialize the population
\medskip

$\bullet\quad$ Find the best fit individual $\widehat y$
\medskip

$\bullet\quad$ Initialize the mutation control
\medskip

\noindent
Main loop
\medskip

$\bullet\quad$ Set $\widehat x=\widehat y$
\medskip

$\bullet\quad$ Apply selection
\medskip

$\bullet\quad$ Apply mutation
\medskip

$\bullet\quad$ Apply crossover
\medskip

$\bullet\quad$ Find the best fit individual $\widehat y$
\medskip

If $\text{fitness}(\widehat x)<\text{fitness}(\widehat y)$ then
\medskip

\qquad Reinitialize the mutation control
\medskip

\qquad Set the population size to two
\medskip

\qquad Initialize the population with two copies of 
$\widehat y$
\medskip

\qquad Goto Main loop
\medskip

If the mutation probability has converged then
\medskip

\qquad Increase the population size
\medskip

\qquad Reinitialize the mutation control
\medskip

\qquad Goto Main loop
\medskip

If $\text{fitness}(\widehat x)=\text{fitness}(\widehat y)$ then
\medskip

\qquad Increase mutation probability
\medskip

\qquad Goto Main loop
\medskip

If $\text{fitness}(\widehat x)>\text{fitness}(\widehat y)$ then
\medskip

\qquad Decrease mutation probability
\medskip

\qquad Reintroduce $\widehat x$ in the population
\medskip

\qquad Set $\widehat y=\widehat x$
\medskip

\qquad Goto Main loop
\vfill
\newpage
\noindent
\bigskip

\noindent
{\bf Acknowledgements:} 
This work was completed during a visit to the mathematics department of
the University of Padova.
I warmly thank Carlo Mariconda and Paolo Dai Pra for their hospitality.

\bibliographystyle{plain}
\bibliography{cga}
\bigskip

\noindent
{Rapha\"el Cerf\\
Universit\'e Paris Sud\\
Math\'ematique, B\^atiment~$425$\\
91405 Orsay Cedex--France\\
rcerf@math.u-psud.fr}

\end{document}